\documentclass[runningheads]{llncs}
\usepackage{graphicx}
\usepackage{amsmath}

\usepackage{booktabs}
\usepackage{subcaption}
\usepackage{multicol}
\usepackage{hyperref}
\newcommand{\mulcol}[1]{\multicolumn{4}{c}{#1}}

\begin{document}
%
% \title{Back to Simplicity: Discussion on Robust Data Augmentation for Multi-organ Segmentation}
% \title{Exploring Robust Data Augmentation for Limited Organ Segmentation Datasets}
\title{Cut to the Mix: Simple Data Augmentation Outperforms Elaborate Ones in Limited Organ Segmentation Datasets}
% \title{Improving Organ Segmentation on Limited Datasets using Inter-Image Data Augmentation}
% \title{Improving Organ Segmentation on Limited Datasets using Robust Data Augmentation}
% \title{Improving Organ Segmentation on Limited Datasets using Innovative Data Augmentation}
% \title{Return to Simplicity: Investigation on Robust Data Augmentation for Organ Segmentation in Limited Dataset}
%
\titlerunning{Data Augmentation for Multi-organ Segmentation}
% If the paper title is too long for the running head, you can set
% an abbreviated paper title here
%
% \author{Anonymous}
\author{Chang Liu\inst{1} \and		%1{Chang, Liu}
Fuxin Fan\inst{1} \and					%2{Fuxin, Fan}
Annette Schwarz\inst{1,2} \and		%3{Annette, Schwarz}
Andreas Maier\inst{1}}					%4{Andreas, Maier}
%1{Chang, Liu}
%2{Fuxin, Fan}
%3{Annette, Schwarz}
%4{Andreas, Maier}

\authorrunning{C. Liu et al.}
% \authorrunning{Anonymous}
% First names are abbreviated in the running head.
% If there are more than two authors, 'et al.' is used.
%
% \institute{Anonymous Institution}
\institute{Pattern Recognition Lab, Department of Computer Science, Friedrich-Alexander-Universität Erlangen-Nürnberg, Martensstr. 3, 91058 Erlangen, Germany \and
Siemens Healthineers, 91027 Forchheim, Germany
}
\maketitle              % typeset the header of the contribution
\begin{abstract}
Multi-organ segmentation is a widely applied clinical routine and automated organ segmentation tools dramatically improve the pipeline of the radiologists. 
Recently, deep learning (DL) based segmentation models have shown the capacity to accomplish such a task.
% The robust training of the segmentation networks requires large amount of image data with manual annotation. 
% However, such large scale segmentation datasets are difficult to acquired for new imaging modalities and the related researches are limited by low amount of existing data, for example for dual energy CT (DECT). 
However, the training of the segmentation networks requires large amount of data with manual annotations, which is a major concern due to the data scarcity from clinic. 
Working with limited data is still common for researches on novel imaging modalities. 
% Applying data augmentation (DA) for training segmentation dataset is a common practice to scale up the generalizability of the limited dataset. 
To enhance the effectiveness of DL models trained with limited data, data augmentation (DA) is a crucial regularization technique.
% Traditional DA strategies are well proven to extend the robustness of the network training.
Traditional DA (TDA) strategies focus on basic intra-image operations, i.e. generating images with different orientations and intensity distributions. In contrast, the inter-image and object-level DA operations are able to create new images from separate individuals. 
However, such DA strategies are not well explored on the task of multi-organ segmentation. 
% In this paper, we conduct experiments on two different datasets using four possible inter-image DA
% Recently, many efforts have been dedicated to the advanced DA strategies in computer vision researches, CutMix for example, to further extend the traditional DA strategies. 
% However, such DA are rarely evaluated for the organ segmentation task because they sometimes destroy the human anatomy of a medical image.
% In this work, we evaluated four object-level DA strategies in order to offer more insight onto the robust DA strategies for organ segmentation task on limited segmentation dataset. 
In this paper, we investigated four possible inter-image DA strategies: CutMix, CarveMix, ObjectAug and AnatoMix, on two organ segmentation datasets. 
The result shows that CutMix, CarveMix and AnatoMix can improve the average dice score by 4.9, 2.0 and 1.9, compared with the state-of-the-art nnUNet without DA strategies. 
These results can be further improved by adding TDA strategies.
% it is revealed in our experiments that the most powerful data augmentation method is the most simple but not anatomy-aware strategy, CutMix. 
It is revealed in our experiments that CutMix is a robust but simple DA strategy to drive up the segmentation performance for multi-organ segmentation, even when CutMix produces intuitively `wrong' images.
% We present our implementation as a DA toolkit for multi-organ segmentation on GitHub for future benchmarks.
Our implementation is publicly available at \href{https://github.com/Rebooorn/mosDAtoolkit}{https://github.com/Rebooorn/mosDAtoolkit} for future benchmarks.
% On the DECT dataset, the CutMix improves the average dice score using nnUNet by 1.7 \% with TDA. On the AMOS dataset, the CutMix improves by 3.4 \% with TDA and 6.4 \% without TDA. 

\end{abstract}
\section{Introduction}
% ideas:
% organ segmentation is routine and automated tool is expected.
% DL offers a working solution and DL relies on robust dataset.
% large-scale datasets are there for common modalities but not so much for new modalities.
% Scaling up the generalizability of the limited dataset will benefit the research of such modalities. DA will be a good option.
% The research on DA in computer vision is widely available. However such DA will not be applied to organ segmentation task, because ...
% In this research, we target at the robust DA to scale up the network performance on limited segmentation dataset. Several recently proposed DA in CV and tumor tasks are re-implemented.
% All re-implementation is now available on GitHub for further benchmarking.

In clinics, multi-organ segmentation is a common routine for radiologists in order to perform a multitude of treatments or therapies, i.e. for the treatment planning for radiation therapy \cite{nikolov2018deep}. 
However, manual delineation of human organs in medical images is a time- and effort-consuming task and automated multi-organ segmentation is thus expected. 
% In additional for the research of CT radiation dose, segmentation of radiation-sensitive organs will contribute to the analysis of radiation risk of each organ.
In recent years, the emerging deep learning (DL) based models have shown strong performance on the task of organ segmentation in some imaging modalities, such as computed tomography (CT) and magnetic resonance scanning (MR) \cite{isensee2021nnu,wasserthal2023totalsegmentator}.
For supervised training routines, the development of a robust DL segmentation model relies on a large-scale image dataset with manual annotation of organs. 
Along with the state-of-the-art organ segmentation models, many large-scale segmentation datasets are publicly available  \cite{ma2021abdomenct,ji2022amos,wasserthal2023totalsegmentator}.
However, for research on novel imaging modalities such as dual energy computed tomography (DECT), it remains difficult to gather enough images.
For preliminary researches, such as to investigate whether certain imaging modalities will benefit the DL models for organ segmentation, working with a limited segmentation dataset is still common \cite{chen2018dect}.
% Recently, the emerging of large-scale organ segmentation datasets accelerates the development and modern performance of multi-organ segmentation tools. 
% On the one hand, the focus of multi-organ segmentation research pushes towards larger datasets with more organs manually annotated. 
% However, on the other hand, the research on novel image modalities still struggle to gather enough data for training robust networks, for example dual energy CT.

It is a reasonable practice to extend the generalizability of limited segmentation dataset using data augmentation (DA), which is a widely known regularization method for DL \cite{Goodfellow2016}. 
Traditional data augmentation (TDA) strategies include spatial shift or scaling, and intensity scaling. 
These are effective for regularizing training processes with more diverse training data. 
Novel approaches, like inter-image and object-level DA strategies are proposed in computer vision and medical imaging research, seeking to reach further than TDA strategies.
Mixup \cite{zhang2018mixup} and CutMix \cite{yun2019cutmix} were first proposed to fuse multiple images from the training dataset for image classification tasks. 
The concept was then adapted into the medical imaging domain, CarveMix \cite{zhang2021carvemix} and selfMix \cite{zhu2022selfmix} are proposed to manipulate the tumor regions within the dataset onto background images for brain and liver tumor segmentation. 
PII \cite{tan2021detecting} is proposed for pathological anomaly detection.
ObjectAug \cite{zhang2021objectaug} introduces object-level data augmentation using the segmentation mask of the components in the image, to apply augmentation for image classification. ClassMix \cite{olsson2021classmix} and ComplexMix \cite{chen2021complexmix} are proposed in researches of autonomous driving to merge the image and the segmentation mask to generate novel street views. 
AnatoMix \cite{liu2024anatomix} is recently proposed in particular for augmentation for multi-organ segmentation task.

To the best of our knowledge, few investigations have been done for inter-image and object-level DA strategies on multi-organ segmentation task. 
In this work, we present our investigation on robust DA strategies for multi-organ segmentation in limited dataset. 
Four established DA strategies have been re-implemented to fit the multi-organ segmentation task: CutMix, ObjectAug, CarveMix and AnatoMix.
% It is commonly presumed that data augmentation methods should create in-distribution data as in the original dataset. 
% % For example, the outputs from the DA strategies should contain the same amount of organs as in the original dataset and, 
% For the image data for organ segmentation, this could lead to the expectation that general characteristics of the human anatomy are preserved, like the number and relative location of organs.
% In other words, two livers and four kidneys in the output images should be avoided.
% % we will hope that the output from the data augmentation machine still have the same amount of organs as in the original datasets. 
% However, it is revealed through our research that such presumption is not true for DL-based networks.

\begin{figure}[h]
    \centering
    \includegraphics[width=0.8\textwidth]{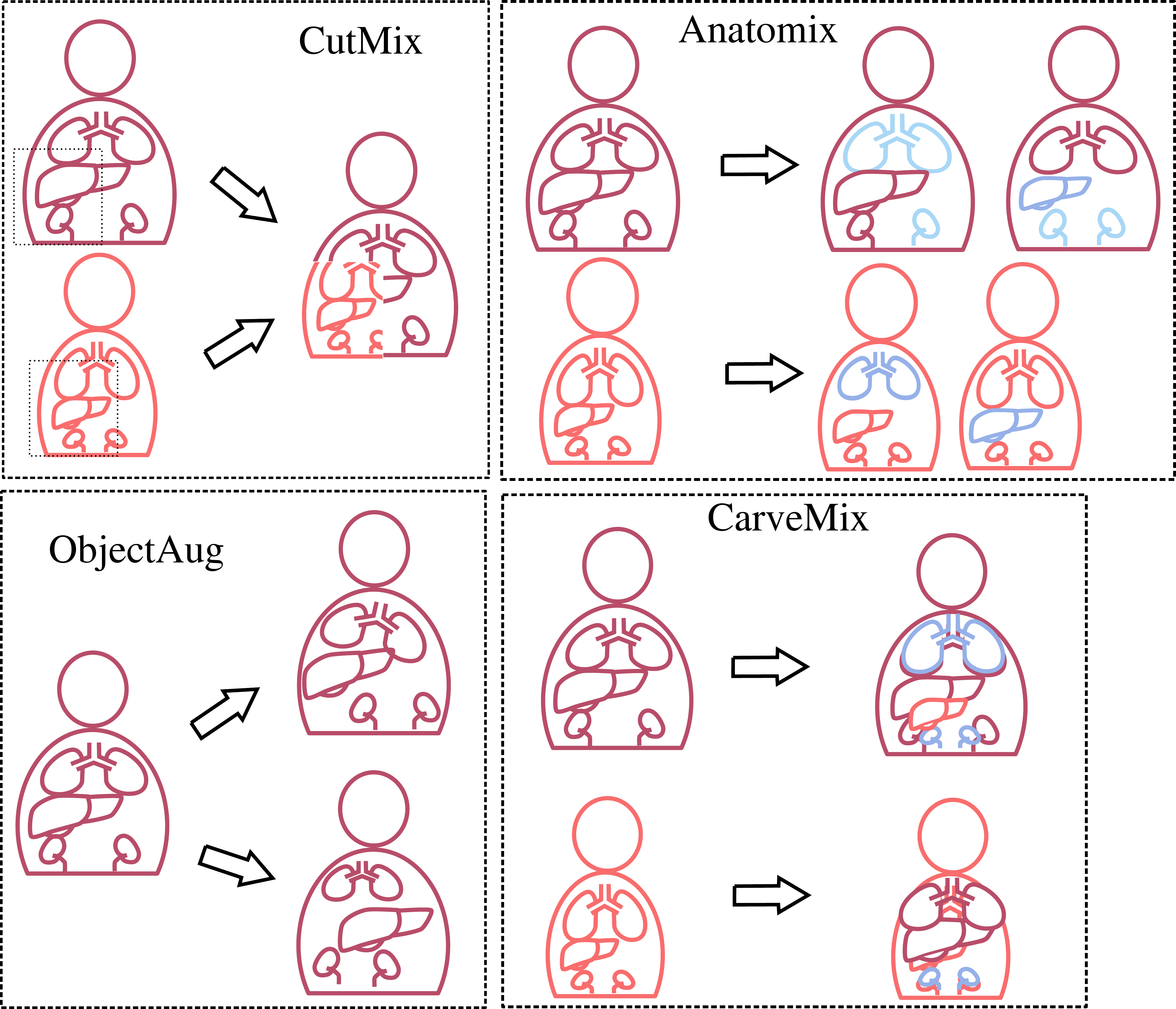}
    \caption{Illustration of the concept of CutMix, ObjectAug, CarveMix and AnatoMix. All DA strategies are originally proposed for either image classification or tumor segmentation task. They are re-implemented for the multi-organ segmentation task and further evaluated in this work.}
    \label{fig:illustration}
\end{figure}

\section{Method}
The aforementioned DA strategies are first re-implemented for multi-organ segmentation tasks and evaluated on two limited organ segmentation tasks, aiming to find a robust DA for multi-organ segmentation. 
nnUNetv2 is applied to train the segmentation networks \cite{isensee2021nnu}.
%, as a reliable training framework to put aside other factors to cause decreased performance and focus only on the DA strategies onto the dataset
% as a strong backbone that the training process is robust and the only factor to the segmentation result is the data augmentation methods.

\subsection{Inter-image and Object-level Data Augmentation}
Assume the segmentation dataset contains $N_d$ cases with \(\{I_i, M_i | 0<i\leq N_d\}\), where $I_i\in R^{D\times W\times H}$ is the volume image and the $M_i\in R^{N_{og}\times D\times W\times H}$ is the manual annotation of $N_{og}$ organs or structures. 
$M_i$ also consists of multiple channels for all organs, so $M_i=\{m_i^j|0<j\leq N_{og}\}$, where $m_i^j\in R^{D\times W\times H}$ is the binary mask of a specific organ in image $I_i$. 
A common operation for object manipulation in images is to overlap region of a selected image, or mask, $I_s$ onto a background image, or mask, $I_b$ using a binary mask $m$. For simplicity, in the following manuscript such fusion operation is formulated as 
\begin{equation}
    I_b \otimes (I_s,b) = I_s \cdot m + I_b \cdot (1-m).
\end{equation}
The four DA strategies are illustrated in Fig. \ref{fig:illustration} and their potential outputs are compared in Table. \ref{tab:augcompare}.

% \begin{table}[h]
%     \centering
%     \caption{Comparison of advanced data augmentation methods for multi-organ segmentation task.}
%     \begin{tabular}{|c|p{2.5cm}|p{2.5cm}|p{2.5cm}|p{2.5cm}|}
%         \hline
%         The Outputs... & have correct Num. of organs?  & have correct Organ Location? & have complete organ regions? & have inpainted area? \\
%         \hline
%         CutMix    & No & No & No & No\\
%         ObjectAug & Yes & No & Yes & Yes\\
%         CarveMix  & No & No & Yes & No\\
%         AnatoMix  & Yes & Yes & Yes & No\\
%         \hline
%     \end{tabular}
%     \label{tab:augcompare}
% \end{table}

\begin{table}[t]
    \centering
    \caption{Comparison of the DA strategies re-implemented to multi-organ segmentation task.}
    \begin{tabular}{|l|c|c|c|c|c|}
    \hline
        Do the DA outputs\dots  & TDA & CutMix & ObjectAug & CarveMix & AnatoMix \\ \hline
        have correct number of organs  & Yes & No & Yes & No & Yes \\
        have correct organ locations   & Yes & No & Yes & No & Yes \\
        cause broken organs            & No & Yes & No & No & No\\
        have artificial voxels         & No  & No & Yes & No & No \\
        \hline
        
    \end{tabular}
    \label{tab:augcompare}
\end{table}

\paragraph{CutMix} CutMix is originally proposed in computer vision research \cite{yun2019cutmix} and two images are fused to a new classification label. 
For the multi-organ segmentation task, a random bounding box mask $m_{bb}$ is created with center at a random location with size following a ratio proportional to the image size following a Beta distribution $\beta(0.5,0.5)$. 
The CutMix is then formulated as:
\begin{align}
    I' & = I_b \otimes (I_s,m_{bb}),\\
    M' & = \{m_b^j \otimes (m_s^j,m_{bb})\},
\end{align}
where $I_b$ is the background image and $I_s$ is a randomly selected image from the source dataset.

\paragraph{ObjectAug} Like CutMix, ObjectAug is also created for the classification task. The concept is to disassemble the components within the image first and then augment each object. A background inpainting model $\theta_b(I, m_{hole})$ is thus needed because the disassemble-recombine process will come with a binary hole mask $m_{hole}$ \cite{liu2018partialinpainting}. 
We implemented random scaling by 10\%, random shift by 5 voxels in all dimension and random rotation of 15$^\circ$ for object-level augmentation, termed as $G^j$. The recombination process loops over each organ. The ObjectAug is then formulated as
\begin{align}
    I^0 &= I'_b, I^{j+1} = I^j \otimes (G^j(I_b),G^j(m^j_b)), \\
    I' & =\theta_b(I^{N_t}, m_{hole}),\\
    M' & = \{G^j(m^j_b)\},
\end{align}
where $I'_b$ is $I_b$ by setting all organ pixels to background. 

\paragraph{CarveMix} CarveMix is proposed to combine the brain tumor with healthy brain region to extend the brain tumor segmentation dataset. For the multi-organ segmentation, CarveMix can be applied to each individual organ and the augmentation is then formulated as 
\begin{align}
    I^0 & = I_b, I^{j+1} = I^j \otimes (I_s, m_s^j), \\
    I' & = I^{N_t}, \\
    M' & = \{m^j_b \otimes (m^j_s, m^j_s)\},
\end{align}
where $I_b$ and $I_s$ are the background image and a randomly selected image from the source dataset.

\paragraph{AnatoMix} CarveMix for multi-organ segmentation will not maintain the human anatomy, in cases of organ location and organ size as can be seen in Fig. \ref{fig:illustration}.
To counter that, AnatoMix contains two steps: augmentation planning and organ transplant. 
First, the sizes of each single organ $m_i^j$ in the dataset are analysed and a organ $m_{i'}^{j'}$ from image $I_{i'}$ with similar size will be matched for each organ $m_i^j$.
Each organ in the background image can then be `replaced' with similar organs in the dataset, shifted by an optimal offset $S_i^j$. The augmentation is formulated as 
\begin{align}
    I^0 & = I_b, I^{j+1} = I^j \otimes (S_i^j(I_{i'}), S_i^j(m_{i'}^{j'})), \\
    I' & = I^{N_t} \\
    M' & = \{m^j_b \otimes (S_i^j(m_{i'}^{j'}), S_i^j(m_{i'}^{j'}))\} 
\end{align}

% \begin{figure}[ht]
%     \centering
%     \includegraphics[width=\textwidth]{pipeline_v1.pdf}
%     \caption{The pipeline of AnatoMix}
%     \label{fig:pipelinev1}
% \end{figure}

% \begin{figure}[ht]
%     \centering
%     \includegraphics[width=\textwidth]{pipeline_v2.pdf}
%     \caption{The pipeline of AnatoMix V2. This architecture is less fun, but it really works better on dice.}
%     \label{fig:pipelinev2}
% \end{figure}

\subsection{Data}
% . One with truncated training dataset but the full test dataset. The other with full training and test dataset.
Two organ segmentation datasets are used for the evaluation: the public abdominal multi-organ segmentation (AMOS) dataset and a private DECT dataset. AMOS dataset contains 300 abdominal CT volumes with segmentation of 16 organs and anatomical structures: spleen (spln), left kidney (lkdy), right kidney (rkdy), gall bladder (gbdr), esophagus (ephs), liver (livr), stomach (stmh), aorta (arta), postcava (pscv), pancreas (pcrs), right adrenal gland (rdrg), left adrenal gland (ldrg), duodenum (ddn), bladder (bldr) and prostate (prst). 
The DECT dataset is collected in the university hospital of Erlangen and manually annotated by a medical student, verified by a medical supervisor. 
The DECT dataset contains 42 CT images with segmentation of 9 abdominal organs: left kidney (lkdy), right kidney (rkdy), liver (livr), spleen (spln), left lung (llng), right lung (rlng), pancreas (pcrs), gall bladder (gbdr) and aorta (arta). 
For AMOS dataset, the training dataset is truncated to have only 20 images, for simulation of a limited dataset, and the full test dataset, i.e. 100 test images, are used.  
For DECT dataset, 20 images are used for training and 22 images for test. 

In addition to the annotated organs, the two datasets also differ in the scanning regions. 
% Because the data are gathered from multiple institutes, the AMOS datasets contains diverse scanned regions. 
The AMOS dataset contains diverse scanning regions, but for DECT dataset the scanning region is almost consistent 
% In contrast for the DECT dataset, the scanned region is very consistent 
because the data comes from one single institute within a same time period.

\subsection{Experimental Setting}
% The previous two datasets are selected to investigate whether the DA strategies rely on the 

For each DA strategy, we investigated the impact by the training dataset, the augmentation multiplier and the compatibility with the TDA strategies. 
For each dataset, we apply each DA strategy to augment 10, 25 and 50 times the size of the original dataset, namely 200, 500 and 1,000 images. 
The cases in the original dataset are not in the augmented training dataset.
Then the nnUNet is trained on each augmented dataset. 
% The utility of automatic experiment planning of nnUNet keeps the robust training of networks across the experiments as much as possible, indicating that the segmentation performance is only influenced by the augmented datasets.
The nnUNet framework is selected because the training dataset is automatically resampled in every training epoch to a fixed number of steps, so that the test performance is controlled as much as possible to only depend on the DA strategies being applied.

The DA strategies are first evaluated with no TDA to focus on the increase of generalizability. 
Also we present the performance of each DA strategy with the optimized TDA of nnUNet, as it is a common practice to combine such DA strategies with TDAs.
The dice scores (dsc) of each organ are used for evaluation and aggregated in two ways: The macro averaged dsc aggregates the dice score of each single organ in each test sample, and the micro averaged dsc aggregates each metric from each organ in each sample, then lead to the globally averaged dsc.
The micro averaged dsc is effective to indicate the general accuracy of prediction and the macro averaged dsc is more sensitive to segmentation of small objects, both are important factors for multi-organ segmentation.
All experiments are done on 4 Nvidia A100 GPU (40G).

\section{Results and Discussion}

\begin{figure}[ht]
    \centering
    \includegraphics[width=0.9\textwidth]{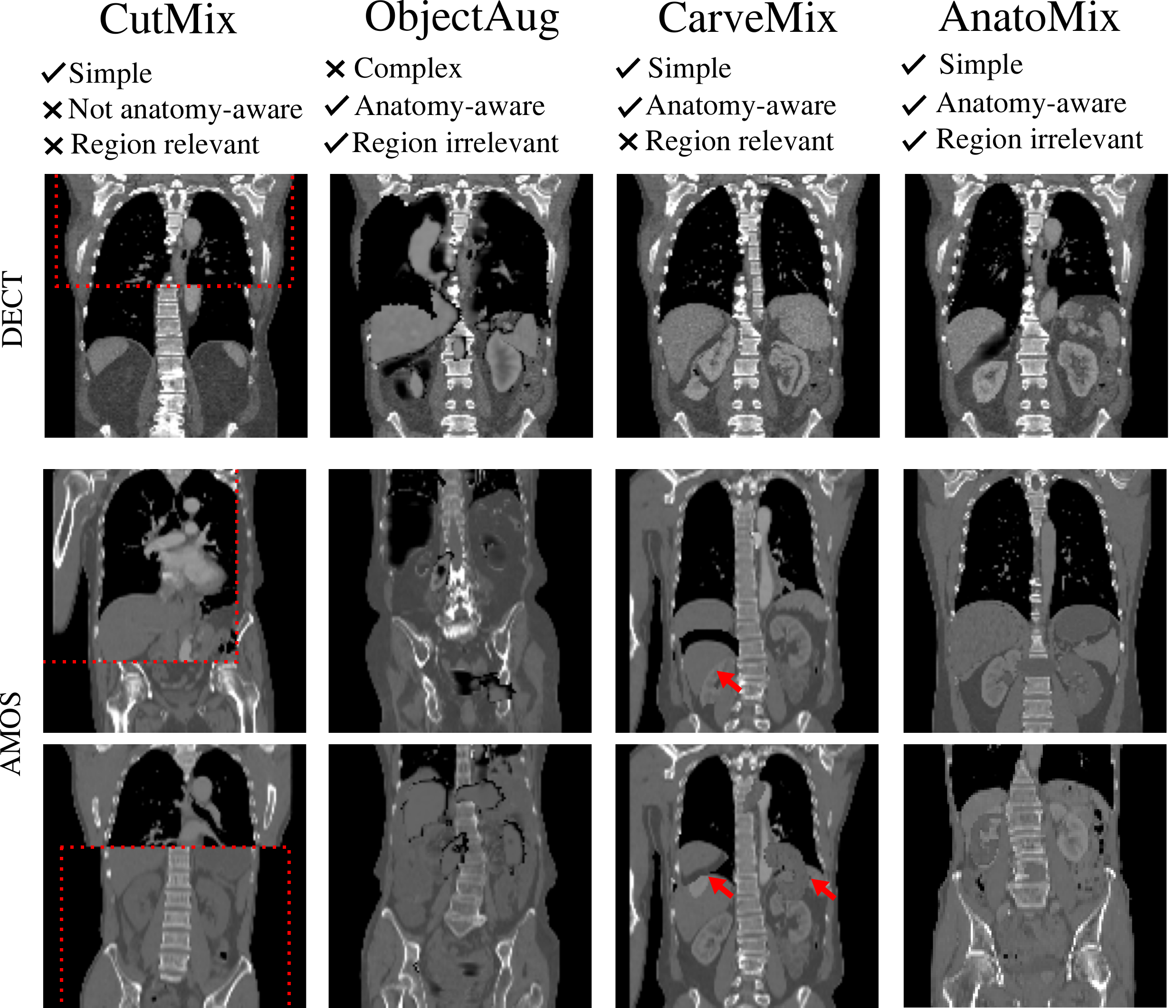}
    \caption{Some example slices of the output volumes using the different DA strategies. The first row shows the outputs from the DECT dataset and the lower rows show the outputs from AMOS dataset. The four DA strategies lead to different compliance with the original human anatomy. Red arrows and dashed lines indicate the abnormal regions.}
    \label{fig:aug_res}
\end{figure}

\begin{table}[h]
    \centering
    \caption{Results on the AMOS dataset. 'Micro' and 'Macro' indicate two aggregation strategies. All models are tested on 100 images and trained on 20$\times$multiplier augmented data. Three multipliers are investigated: $\times$10, $\times$25, $\times$50.
    }
    \label{tab:amos:res}
    \begin{tabular}{lcccccccc} \toprule
              & \mulcol{Micro} & \mulcol{Macro} \\ \cmidrule(lr){2-5}\cmidrule(lr){6-9}
                 & $\times$1 & $\times$10 & $\times$25 & $\times$50 & $\times$1 & $\times$10 & $\times$25 & $\times$50 \\ \midrule 

    NoTDA         & 88.1 & -- & -- & -- & 75.9 & -- & -- & --        \\
    +CutMix      & -- & 89.7 & 90.3 & \textbf{90.7} & -- & 78.7 & 79.9  & \textbf{80.8}            \\ 
    +ObjectAug   & -- & 46.7 & 50.4 & 44.4 & --& 7.4 & 8.9 & 8.1  \\ 
    +CarveMix    & -- & 88.6 & 89.3 & 89.4 & --& 77.9 & 77.6 & 77.5    \\ 
    +AnatoMix    & -- & 88.8 & 88.9 & 88.8 & --& 77.1 & 77.8  & 77.7   \\ \midrule 
    TDA          & 88.1 & -- & -- & -- & 78.2 & -- & -- & --          \\
    +CutMix      & -- & 90.6 & 91.0 & \textbf{91.1} & -- & 82.3  & 82.8  & \textbf{83.0}          \\ 
    +ObjectAug   & -- & 66.0 & 64.4 & 64.3 & -- & 25.7 & 22.7 & 22.6  \\ 
    +CarveMix    & -- & 89.2 & 89.7 & 89.6 & -- & 80.0  & 80.7  & 80.2    \\ 
    +AnatoMix    & -- & 89.9 & 89.9 & 89.7 & -- & 81.5  & 81.4  & 80.8   \\  
     \bottomrule
         
    \end{tabular}
\end{table}

\begin{table}[h]
    \centering
    \caption{Results from experiments on the DECT dataset. All models are tested on 20 images and trained on 20$\times$ multiplier augmented data. Three multipliers are investigated: $\times$10, $\times$25, $\times$50
    }
    \label{tab:dect:res}
    \begin{tabular}{lcccccccc} \toprule
              & \mulcol{Micro} & \mulcol{Macro} \\ \cmidrule(lr){2-5}\cmidrule(lr){6-9}
                 & $\times$1 & $\times$10 & $\times$25 & $\times$50 & $\times$1 & $\times$10 & $\times$25 & $\times$50 \\ \midrule 
              
    NoTDA         & 96.9 & -- & -- & -- & 87.7 & -- & -- & --        \\
    +CutMix      & -- & \textbf{97.1} & 97.0 & 97.0 & -- & \textbf{90.8} & 89.1 & 89.3             \\ 
    +ObjectAug   & -- & 76.1 & 77.9 & 76.9 & -- & 26.4 & 27.5 & 25.7  \\ 
    +CarveMix    & -- & 96.8 & 96.7 & 96.7 & -- & 90.5 & 89.1 & 89.0    \\ 
    +AnatoMix    & -- & 96.9 & 96.9 & 96.9 & -- & 90.1 & 89.3 & 89.2  \\  \midrule
    TDA          & 96.8 & -- & -- & -- & 89.1 & -- & -- & --          \\
    +CutMix      & -- & 97.0 & \textbf{97.1} & 97.0 & --& 90.6 & 90.4 & 90.7          \\ 
    +ObjectAug   & -- & 90.7 & 91.0 & 89.5 & --& 60.5 & 63.3 & 62.5  \\ 
    +CarveMix    & -- & 96.9 & 96.8 & 96.9 & --& 90.7 & 90.1 & 90.3     \\ 
    +AnatoMix    & -- & 97.0 & 97.0 & 97.0 & --& 90.5 & 90.9 & \textbf{90.9}   \\  
     \bottomrule
         
    \end{tabular}
\end{table}

% For limited segmentation dataset (LSD), most object-level DA can effectively improve the baseline methods.
% With TDA, most DA can still improve the segmentation performance.
% Supervisingly: CutMix contributes most improvements.

The output of the aformentioned DA strategies are shown in Fig. \ref{fig:aug_res}. 
It can be observed that different DA strategies lead to different consistency with the original dataset.
AnatoMix can produce the CT volumes with correct organ location and similar organ size. 
In contrast, CutMix and CarveMix rely on the similarity of both input images.
When the scanning regions are greatly different, i.e. in the AMOS dataset, CutMix and CarveMix will disturb the human anatomy. For example the output volumes may have four kidneys and two livers or the upper body region will be in the lower body, as indicated by the red arrows and dashed lines in Fig. \ref{fig:aug_res}.
On the same device, CutMix takes on average 0.3s for one output image, while it takes 15.7s for CarveMix, 20.9s for AnatoMix and 40.4s for ObjectAug. Because CutMix only combines two images by region-of-interest (ROI), it is much faster than other methods using slow operations, like background in-painting or object rotation.
% CutMix also mixes the body regions, for example the organs lower body region can be in the upper body.

The segmentation results on AMOS and DECT dataset are shown in Table. \ref{tab:amos:res} and Table. \ref{tab:dect:res}, respectively.
The detailed organ-wise results can be found in the supplementary materials.
On AMOS dataset with no TDA, the applied DA strategies, except for ObjectAug, can lead to increased segmentation performance.
In particular, CutMix can improve the micro averaged dsc the most by 2.6 without TDA, followed by CarveMix by 1.3 and AnatoMix by 0.8. Regarding macro averaged dsc, CutMix, CarveMix and AnatoMix each improves by 4.9, 2.0 and 1.9. 
Together with TDA strategies, CutMix leads to improvement of micro dsc by 3.0, followed by AnatoMix by 1.8 and CarveMix by 1.6. CutMix, CarveMix and AnatoMix each lead to the improvement of macro averaged dsc by 4.8, 2.5 and 3.2.
With or without TDA, increasing the augmentation multiplier can increase the micro and macro averaged dsc on the AMOS dataset for CutMix, but not for CarveMix and AnatoMix. The increase of macro averaged dice is higher than that for micro averaged dsc, indicating the segmentation of small organs is improved. Moreover, the increase of macro averaged dsc by CutMix is higher than that by the optimized TDA from nnUNet, and both increases are additive, leading to a joint increase of 7.0 compared with no TDA.
% Regarding the augmentation multiplier, on AMOS dataset, increasing augmentation multiplier can lead to higher increase of micro and macro averaged dice for CutMix, while not for CarveMix and AnatoMix.
% On all experiments, regardless using TDA or not and for all augmentation multipliers, objectaug all dramatically degrade the segmentation performance.

On the DECT dataset, the baseline performance without any DA already push towards quite high dsc, potentially because the dual channel inputs lower the difficulty of segmentation. 
Still, except for ObjectAug the DA strategies can slightly increase the segmentation performance. 
In particular without TDA, all DA strategies lead to no improvement in micro averaged dsc. In contrast, CutMix leads to improvements of 3.1 in macro averaged dsc, followed by CarveMix and AnatoMix by 2.8  and 2.4. Similarly with TDA applied,  CutMix, CarveMix and AnatoMix each leads to increase by 1.6, 1.6 and 1.8. while no improvement is observed in micro averaged dsc.
Different from the results on the AMOS dataset, increasing augmentation multiplier will not increase the micro or macro dsc and can even decrease the macro averaged dsc for CutMix, CarveMix and AnatoMix when no DA applied. 
Nevertheless, the difference between the highest and lowest macro averaged dsc is close, which indicates that DA strategies will lead to less improvement when the segmentation result is already high enough with limited datasets.

% Different from the results on the AMOS dataset, increasing augmentation multiplier will not increase the micro or macro dsc for all DA strategies. Also ObjectAug failed on the DECT dataset.

% To the best of our knowledge, few investigations have been done for advanced DA strategies on multi-organ segmentation task. 
It is commonly presumed that data augmentation methods should create in-distribution data as in the original dataset. 
% For example, the outputs from the DA strategies should contain the same amount of organs as in the original dataset.
For the image data for organ segmentation, this could lead to the expectation that general characteristics of the human anatomy are preserved, like the number and relative location of organs.
In other words, two livers and four kidneys in the output images should be avoided.
% we will hope that the output from the data augmentation machine still have the same amount of organs as in the original datasets. 
However, it is revealed through our research that such presumption is not always true for DL-based networks. 
% We present the re-implementation of the mentioned DA strategies as a DA toolkit for multi-organ segmentation on GitHub for future benchmarks.

\section{Conclusion}
% Conclusion: 
%       It is found that three DAs can be beneficial and one not. 
%       From metrics, CutMix can be a simple but robust DA for organ segmentation [for limited case]
% Discussion: 
%       The final results are superising, and against intuition.
%       Anatomix can be used for other senarios.
From our experiments, it can be concluded that the CutMix, CarveMix and AnatoMix can effectively enhance the limited segmentation datasets. 
In practice, it is a working strategy to combine such DA strategies with TDA to yield a joint improvement of the segmentation performance.
Surprisingly, from the metric results in our experiment and the complexity of implementation, the CutMix is the best DA strategy for limited multi-organ segmentation datasets. 

% From our results, it is out of initial expectation that CutMix will be the simplest but, at the same time, the most robust DA strategy, rather than the most intuitive AnatoMix. 
% It is presumed that combining training data for DA is a working strategy and CutMix fuses the local information more efficiently than other DA strategies, which could explain its superiority.

% In addition, this research raises a general research question that the DA strategies are not necessary to be task-specific. 
% It is straight-forward to put task-specific constraints when new DA strategies are designed, usually more complexity are added for such constraints, such as human anatomy. However, such intuitions may not be the most efficient solution. 

\begin{credits}
\subsubsection{\ackname} The authors gratefully acknowledge the HPC resources provided by the Erlangen National High Performance Computing Center (NHR@FAU) of the Friedrich-Alexander-Universität Erlangen-Nürnberg (FAU). The hardware is funded by the German Research Foundation (DFG).

\subsubsection{\discintname}
The authors have no competing interests to declare that are relevant to the content of this article.
\end{credits}

\clearpage

\bibliographystyle{splncs04}
\bibliography{citation}

\end{document}